\documentclass[11pt,a4paper]{article}
\usepackage[utf8]{inputenc}
\usepackage[T1]{fontenc}
\usepackage{amsmath,amssymb}
\usepackage{booktabs}
\usepackage{graphicx}
\usepackage{hyperref}
\usepackage{natbib}
\usepackage[margin=2.5cm]{geometry}
\usepackage{xcolor}
\usepackage{float}
\usepackage{multirow}
\usepackage{caption}
\usepackage{tikz}
\usetikzlibrary{arrows.meta,positioning,decorations.pathreplacing,calc}

\hypersetup{
    colorlinks=true,
    linkcolor=blue!60!black,
    citecolor=blue!60!black,
    urlcolor=blue!60!black
}

\title{Evidence of Layered Positional and Directional Constraints\\in the Voynich Manuscript: Implications for Cipher-Like Structure}

\author{
        Christophe Parisel\thanks{Email: ch.parisel@gmail.com}
}

\date{\today}

\begin{document}

\maketitle

\begin{abstract}
The Voynich Manuscript (VMS) exhibits a script of uncertain origin whose grapheme sequences have resisted linguistic analysis. We present a systematic analysis of its grapheme sequences, revealing two complementary structural layers: a character-level right-to-left optimization in word-internal sequences and a left-to-right dependency at word boundaries, a directional dissociation not observed in any of our four comparison languages (English, French, Hebrew, Arabic). Mutual information decomposition shows that 97\% of cross-boundary MI resides in specific within-class grapheme transitions rather than in class-level labels, and a Markov simulation demonstrates that the directional dissociation is reproducible from surface word-level statistics without reference to positional classes or generative mechanisms. Positional analysis identifies Zipfian boundary distributions and an 80.6\% end-class$\to$start-class transition rate, both quantitatively distinct from the comparison languages.

We further evaluate two classes of structured generator against a four-signature joint criterion calibrated separately for each of the two Currier dialects (A and B) identified in the manuscript. A parametric slot-based generator and a Cardan grille implementing Rugg's (2004) gibberish hypothesis are tested across their full parameter spaces. Neither class reproduces all four signatures simultaneously for either dialect. The slot-based generator cannot achieve high cross-boundary MI and the E$\to$S rate together; the grille cannot achieve the E$\to$S rate and Zipfian shape together. These are structurally distinct failure modes arising from mechanistically different generators. While these results do not rule out generator classes we have not tested, they provide the first quantitative benchmarks (dialect-specific and jointly evaluated) against which any future generative or cryptanalytic model of the VMS can be evaluated, and they suggest that the VMS exhibits cipher-like structural constraints that are difficult to reproduce from simple positional or frequency-based mechanisms alone.
\end{abstract}

\section{Introduction}

The Voynich Manuscript (MS 408, Yale University Beinecke Library; hereafter VMS) remains one of the most studied undeciphered texts in history. Three competing hypotheses dominate scholarly discourse: that the VMS is meaningless gibberish \citep{rugg2004, timm2020, gaskell2022}, that it represents a natural or constructed language \citep{bowern2021}, or that it is a ciphertext of a known language such as Latin or Italian \citep{dImperio1978, greshko2025}.

A fundamental yet underexplored question concerns the \emph{directionality} of the Voynich script. While the manuscript was almost certainly \emph{written} left-to-right (LTR), based on ink directionality and stroke analysis, whether the underlying text should be \emph{read} LTR or right-to-left (RTL) has received little quantitative treatment. In prior work \citep{parisel2025}, we introduced a language-agnostic n-gram perplexity asymmetry method that revealed consistent RTL optimization in the VMS character stream.

That same work also uncovered an anomaly at word boundaries. \citet{ashraf2018} showed that natural languages universally exhibit asymmetric grapheme distributions at word boundaries, a property exploitable for directionality detection using Gini and Shannon entropy measures. However, when we applied these boundary metrics to the VMS, we found that they were unsuitable: the distribution of word-initial and word-final graphemes in the VMS follows a Zipfian curve, in stark contrast to the plateau-shaped distributions observed in the comparison languages. This distributional difference (which we call the \emph{Zipfian boundary effect}) means that conventional boundary-based directionality metrics cannot be meaningfully benchmarked against the comparison languages for the VMS.

Motivated by these observations, we developed two new analytical methods that work \emph{across} word boundaries rather than \emph{at} them, circumventing the Zipfian boundary effect while probing the positional properties of VMS words. A Markov simulation establishes that the directional dissociation is reproducible from surface word-level statistics alone, without reference to positional classes or any generative hypothesis. This is a clarifying result, not a deflating one: it rules out the directional dissociation as an independent diagnostic and redirects analytical attention to the positional properties that are \emph{not} reproducible by surface resampling. It is those properties, and their joint behaviour under generative testing, that form the main contribution of this paper.

With the directional dissociation set aside as a surface-level consequence of VMS word structure, we ask what structural properties of the VMS cannot be similarly explained. The answer is a set of four positional signatures: boundary concentration, bilateral positional extremity, cross-boundary mutual information, and Zipfian boundary distributions. We test whether these can be jointly reproduced by two classes of structured generator, a parametric slot-based generator and a Cardan grille generator \citep{rugg2004}, each tested across its full parameter space.

An important analytical refinement in this work is the separate evaluation of the two \emph{Currier dialects}. \citet{currier1976} identified two statistically distinct ``languages'' (conventionally labelled A and B) within the VMS, associated with different sections of the manuscript and differing in word-frequency distributions, character usage, and line-level statistics. Rather than treating the VMS as a single homogeneous corpus, we calibrate the four-signature criterion independently for each dialect and require generators to pass against the appropriate dialect-specific reference. This separation reveals that the two dialects, while sharing the same broad structural architecture, impose quantitatively different constraints: a generator that passes Currier~B may fail Currier~A, or vice versa.

Neither generator class achieves all four signatures simultaneously for either dialect; the failure modes are mechanistically distinct and robust across parameter variation, though they do not rule out generator classes we have not tested.

We emphasise at the outset that our comparison set comprises only four languages. This is sufficient to establish that the VMS differs from \emph{these four languages} on the metrics reported, but is far too small to support claims about natural language in general. Throughout this paper, statements about the VMS being ``different'' or ``exceeding'' baselines refer exclusively to the four-language comparison set, not to natural language as a class.

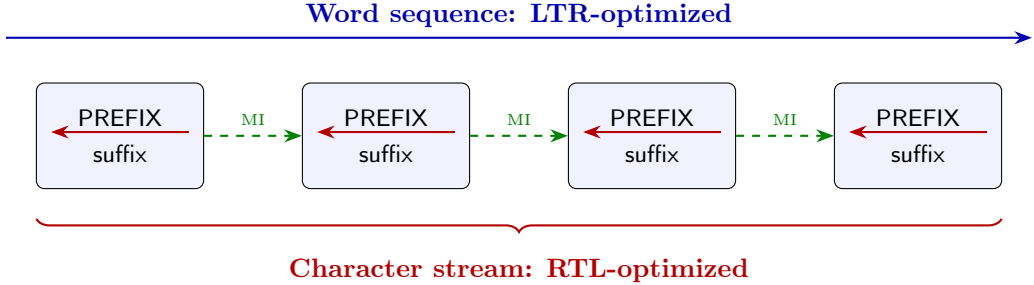
\begin{figure}[H]
\centering
\begin{tikzpicture}[
    word/.style={draw, rounded corners=3pt, minimum width=2.2cm, minimum height=1.4cm, align=center, fill=blue!5},
    >=Stealth
]
\node[word] (w1) at (0,0) {\footnotesize\textsf{PREFIX}\\\footnotesize\textsf{suffix}};
\node[word] (w2) at (3.5,0) {\footnotesize\textsf{PREFIX}\\\footnotesize\textsf{suffix}};
\node[word] (w3) at (7,0) {\footnotesize\textsf{PREFIX}\\\footnotesize\textsf{suffix}};
\node[word] (w4) at (10.5,0) {\footnotesize\textsf{PREFIX}\\\footnotesize\textsf{suffix}};

\draw[->, thick, blue!70!black] (-1.5,1.3) -- (12,1.3);
\node[above, blue!70!black, font=\small\bfseries] at (5.25,1.3) {Word sequence: LTR-optimized};

\foreach \w in {w1,w2,w3,w4} {
    \draw[->, thick, red!70!black] ($(\w.east)-(0.2,-0.05)$) -- ($(\w.west)+(0.2,0.05)$);
}

\draw[decorate, decoration={brace, amplitude=5pt, mirror}, thick, red!70!black]
    (-1.1,-1.1) -- (11.6,-1.1);
\node[below, red!70!black, font=\small\bfseries] at (5.25,-1.5) {Character stream: RTL-optimized};

\draw[->, thick, green!50!black, dashed] (w1.east) -- (w2.west)
    node[midway, above, font=\tiny, green!50!black] {MI};
\draw[->, thick, green!50!black, dashed] (w2.east) -- (w3.west)
    node[midway, above, font=\tiny, green!50!black] {MI};
\draw[->, thick, green!50!black, dashed] (w3.east) -- (w4.west)
    node[midway, above, font=\tiny, green!50!black] {MI};

\end{tikzpicture}
\caption{The two-layer directional profile of the Voynich Manuscript. Words are composed internally of prefix-class and suffix-class graphemes. The character stream is RTL-optimized (red arrows), while word-to-word transitions are LTR-optimized (blue arrow). Cross-boundary mutual information (green, dashed) flows between suffix graphemes of one word and prefix graphemes of the next. A word-level Markov simulation (Section~\ref{sec:markov}) establishes that this two-layer pattern is a surface-level consequence of VMS word-internal structure and local transition probabilities, clarifying that the directional dissociation itself is not diagnostic. The analysis therefore focuses on the positional signatures that are not reducible to surface statistics.}
\label{fig:twolayer}
\end{figure}

Figure~\ref{fig:twolayer} illustrates the two-layer directional profile.

\section{Background}

\subsection{Perplexity-based directionality}

In \citet{parisel2025}, we defined a directional asymmetry measure
\begin{equation}
\Delta = X_{\text{LTR}} - X_{\text{RTL}}
\end{equation}
where $X$ is the average cross-entropy per token under an $n$-gram model. Positive $\Delta$ indicates RTL optimization (lower RTL perplexity); negative $\Delta$ indicates LTR optimization. Applied to the VMS (RF1b-e EVA transcription), we found consistently positive $\Delta$ with tight bootstrap confidence intervals across $n=2$ to $n=4$.

\subsection{The Zipfian boundary anomaly}

\citet{ashraf2018} demonstrated that natural languages universally exhibit asymmetric grapheme distributions at word boundaries, with word-final positions being more constrained than word-initial ones. \citet{winstead2024} extended this finding across a large multilingual corpus, combining Gini index and Shannon entropy into a directional score. However, these boundary metrics assume that word-initial and word-final grapheme frequency distributions follow the plateau-shaped curves characteristic of natural languages.

We showed that the VMS violates this assumption: its boundary grapheme distributions follow a Zipfian (power-law) curve rather than a plateau. This makes Gini and entropy measures unreliable for benchmarking VMS directionality against the comparison languages, as the metrics are sensitive to distributional shape, not just asymmetry. This finding motivated the development of the cross-boundary methods presented in this paper.

\subsection{Positional slot structure}

Several researchers have proposed that VMS words are built from ordered positional slots. \citet{zattera2022} proposed a 12-slot positional structure for Voynich words. \citet{greshko2025} adapted this into a two-class system of type-1 (prefix) and type-2 (suffix) affixes with disjoint glyph sets as part of the Naibbe cipher model. Our analysis does not depend on any specific slot model or cipher hypothesis; we use the concept of positional structure purely as a descriptive framework for characterising the observed distributional properties of VMS graphemes.

\subsection{The Currier dialects}\label{sec:currier}

\citet{currier1976} identified two statistically distinct ``languages'' within the VMS, conventionally labelled Currier~A and Currier~B, based on differences in character frequency, word-frequency distributions, and overall statistical texture. Currier~A predominates in the herbal and pharmaceutical sections; Currier~B predominates in the biological and astronomical sections. The two dialects share the same grapheme inventory and broad positional architecture but differ quantitatively in several measurable properties.

Most prior quantitative analyses of the VMS treat the manuscript as a single corpus. However, since the four-signature criterion evaluates statistical properties that may differ between dialects: particularly MI (which reflects word-to-word sequential dependencies that could vary with section-specific vocabulary) and boundary distribution shape. So we calibrate the criterion separately for each dialect. This ensures that generators are evaluated against the correct reference and prevents a generator from passing on aggregate statistics that neither dialect actually exhibits.

\section{Methods}

\subsection{Data}

We use five corpora:
\begin{itemize}
\item \textbf{Voynich:} RF1b-e EVA transcription \citep{zandbergen2025}, 37,016 words across 5,820 sentences. For dialect-specific analyses, folios are assigned to Currier~A or Currier~B following the standard folio-language map \citep{currier1976}. Currier~A comprises approximately 9,900 words across 108 folios; Currier~B comprises approximately 19,600 words across the remaining folios.
\item \textbf{English (LTR):} Melville's \emph{Moby Dick} \citep{melville}.
\item \textbf{French (LTR):} Dumas' \emph{Le Comte de Monte Cristo} \citep{dumas}.
\item \textbf{Hebrew (RTL):} SVLM Hebrew Wikipedia Corpus \citep{svlm}.
\item \textbf{Arabic (RTL):} The Big Arabic Corpus \citep{arabic}.
\end{itemize}

These four languages were chosen to include both LTR and RTL scripts and both European and Semitic language families. This is a convenience sample, not a typologically representative one; it excludes agglutinative, polysynthetic, and tonal languages, among others. All comparative statements in this paper are limited to these four languages.

EVA graphemes are tokenized using a longest-match greedy algorithm against the STA1 grapheme inventory. Natural language corpora are tokenized at the character level. For Hebrew and Arabic, which are stored in logical (LTR memory) order, we apply a visual transformation (reversing word order within sentences) before cross-boundary analysis so that the ``forward'' direction corresponds to actual reading order.

\subsection{Cross-boundary directionality test}

For each pair of consecutive words $(w_i, w_{i+1})$ in a sentence, we extract the \emph{boundary transition}: a pair consisting of the last $n$ graphemes of $w_i$ (the \emph{condition}) and the first $n$ graphemes of $w_{i+1}$ (the \emph{target}). We compute the conditional entropy
\begin{equation}
H(\text{target} \mid \text{condition}) = -\sum_{c,t} P(c,t) \log_2 P(t \mid c)
\end{equation}
and the mutual information
\begin{equation}
\text{MI}(\text{condition}; \text{target}) = H(\text{target}) - H(\text{target} \mid \text{condition})
\end{equation}
for both forward (original word order) and backward (reversed word order) directions. Word-internal grapheme order is never modified; only the sequence of words varies between conditions.

The directional asymmetry is
\begin{equation}
\Delta_{\text{CB}} = H_{\text{fwd}} - H_{\text{bwd}}
\end{equation}
where positive values indicate RTL optimization and negative values indicate LTR optimization. \textbf{Weighting note.} $\Delta_{\text{CB}}$ is computed from corpus-wide frequency counts and is therefore implicitly weighted by the number of boundary transitions per sentence. Confidence intervals are computed via paired bootstrap resampling over sentences ($B=500$, $\alpha=0.05$), which resamples sentences with equal weight regardless of length. When these disagree in sign, the n-gram order is treated as inconclusive. A shuffle control (random permutation of word order within sentences) verifies that detected signals reflect genuine sequential structure.

\subsection{Positional diagnostic}

We classify each grapheme as \emph{start-preferring}, \emph{end-preferring}, or \emph{ambiguous} based on a 2:1 ratio threshold of word-initial versus word-final occurrence counts. The \emph{polarization index} is the fraction of graphemes that are clearly start- or end-preferring.

We then decompose cross-boundary mutual information into two components:
\begin{equation}
\text{MI}_{\text{total}} = \text{MI}_{\text{class}} + \text{MI}_{\text{within}}
\end{equation}
where $\text{MI}_{\text{class}}$ measures information carried by the positional class labels alone, and $\text{MI}_{\text{within}}$ measures the residual information carried by specific grapheme identities within their positional class.

We apply word-order shuffling to both components to distinguish structural (order-independent) from sequential (order-dependent) contributions.

\subsection{Markov simulation}\label{sec:markov_methods}

To test whether the observed directional combination requires explanation beyond surface word-level statistics, we train a word-level Markov chain on the VMS and generate synthetic corpora:

\begin{enumerate}
\item A word-level Markov chain of order $k$ (tested at $k=1$ and $k=2$) is trained on VMS word sequences, with \texttt{<BOS>} and \texttt{<EOS>} tokens marking sentence boundaries.
\item Synthetic sentences are generated by sampling from the chain, with sentence lengths drawn from the real VMS length distribution. Only words that pass EVA tokenization are retained.
\item Both diagnostics (character-level $\Delta_{\text{char}}$ and cross-boundary $\Delta_{\text{CB}}$) are computed on each synthetic corpus.
\item The procedure is repeated for 10 independent runs at each Markov order.
\end{enumerate}

The simulation preserves VMS word-internal structure (because it samples real VMS words) and approximate word-to-word transition probabilities, but has no knowledge of grapheme-level positional classes or any generative model. If the directional combination is reproducible under these conditions, it is a consequence of surface statistics rather than evidence for a specific generative mechanism.

\subsection{Generative model testing}\label{sec:gen_methods}

To test whether the four positional signatures can be jointly reproduced by structured generators, we implement two distinct generator classes and evaluate each against all four signatures simultaneously.

\paragraph{Dialect-specific four-signature evaluation.}
The four-signature criterion is calibrated separately for Currier~A and Currier~B. For each dialect, a reference profile is computed from the real VMS text of that dialect using 10 evaluation chunks with confidence intervals. A generated corpus passes the joint profile for a given dialect if it simultaneously satisfies: (Sig1) E$\to$S rate within the dialect-specific adaptive range; (Sig2) bilateral positional extremity matching the dialect reference; (Sig3) cross-boundary MI within the dialect-specific adaptive range; and (Sig4) boundary distribution shape matching the dialect reference.

Table~\ref{tab:dialect_reference} presents the calibrated reference values and adaptive thresholds for each dialect.

\begin{table}[H]
\centering
\caption{Dialect-specific reference values and adaptive thresholds for the four-signature criterion. Reference values are computed from the calibration pipeline for each dialect. Adaptive ranges define the passing criterion for each signature.}
\label{tab:dialect_reference}
\begin{tabular}{p{3.5cm}p{3.5cm}p{3.5cm}p{3.5cm}}
\toprule
Signature & Measure & Currier A reference & Currier B reference \\
\midrule
\textbf{Sig1:} E$\to$S & End$\to$Start \% & 71.0\% & 64.3\% \\
 & Adaptive range & [56\%, 86\%] & [50\%, 79\%] \\
\midrule
\textbf{Sig2:} Bilateral & Start \& end extremity & Yes (Se=3, Ee=3) & Yes (Se=3, Ee=7) \\
 & Criterion & Must have bilateral & Must have bilateral \\
\midrule
\textbf{Sig3:} MI & Cross-boundary MI & 0.5856 & 0.4980 \\
 & Adaptive range & $\geq 0.29$ & $\geq 0.25$ \\
\midrule
\textbf{Sig4:} Shape & Distribution type & Zipfian ($R^2$=0.863, CV=1.45) & Intermediate ($R^2$=0.805, CV=1.53) \\
 & Criterion & Non-Plateau & Non-Plateau \\
\midrule
 & Corpus size & 9,892 words & 19,640 words \\
\bottomrule
\end{tabular}
\end{table}

Several dialect differences are noteworthy. Currier~A has a higher E$\to$S rate (71.0\% vs.\ 64.1\%), a higher MI (0.586 vs.\ 0.498), and a Zipfian boundary distribution shape ($R^2 = 0.863$, CV = 1.45), while Currier~B has an Intermediate shape ($R^2 = 0.805$, CV = 1.53). These differences mean that a generator tuned to pass one dialect may fail the other: the MI threshold is higher for Currier~A ($\geq 0.29$ vs.\ $\geq 0.25$ for~B), while the E$\to$S range is wider for Currier~B ([50\%, 79\%] vs.\ [56\%, 86\%]). Both dialects require bilateral extremity and non-Plateau shape, but the specific shape (Zipfian vs.\ Intermediate) differs.

All metrics are computed using the same pipeline as the main analysis. Each configuration is run 20 times with different random seeds; results are reported as bootstrapped means with 95\% confidence intervals. Generators are evaluated against both dialect references and must specify which dialect they target.

\paragraph{Slot-based generator.}
A parametric generator builds synthetic corpora from ordered positional slots, mirroring the structural hypothesis that VMS words are composed from prefix-class and suffix-class graphemes. The generator constructs a lexicon of approximately 1,000 cipher words by sampling from prefix and suffix slot pools, assigns Zipfian frequency weights, and samples word sequences via a sparse Markov chain with tunable boundary-pair preferences. Key parameters include: bridge zone size (graphemes shared between prefix and suffix terminal slots, controlling E$\to$S); Markov sparsity (top-$k$ successors per source word, controlling MI); and Zipf exponent (concentration of within-slot frequency, controlling boundary distribution shape). We test 12 ablation conditions (including single-pool, random-order, overlapping-pool, and near-miss variants) and seven sensitivity sweeps (S1 pool overlap, S2 slot count, S3 Zipf exponent, S4 vocabulary size, S5 boundary pair strength, S6 bridge zone width, S7 Markov top-$k$).

\paragraph{Cardan grille generator.}
A second generator implements the Cardan grille model proposed by \citet{rugg2004} as an alternative to linguistic hypotheses. A grille table of dimensions $(\text{rows} \times \text{cols})$ is filled with graphemes drawn from column-specific pools; a grille (a binary mask of $n_\text{holes}$ column positions) is applied to a selected row, and the non-blank graphemes at hole positions form one word. We implement four word-generation modes: \textsc{random} (Rugg's original model: each word uses an independently drawn grille position), \textsc{shift} (grille advances one column after each word), \textsc{rotate} (grille cycles through all unique rotations), and \textsc{sequential} (words are generated by reading successive rows of a table whose column distributions are learned from a real corpus rather than pre-specified pools).

We distinguish explicitly between configurations that impose pre-separated prefix/suffix pools (\emph{circular on Sig1}) and \emph{honest} configurations where no positional pool structure is imposed. Honest variants include: a uniform-pool table with large row count (establishing the column-skew-free baseline), a blank-gradient table (testing whether density variation alone induces E$\to$S), learned-column-distribution tables built from English text and random grapheme sequences (testing whether real corpus structure transmits through the grille), and row-sequential traversal of a 201,723-word English corpus at four jump probabilities $p \in \{0.00, 0.05, 0.10, 0.30\}$ (testing whether sequential scribe behaviour generates MI). Five sensitivity sweeps cover blank probability, hole count, table columns, column skew, and table rows. 

\subsection{Reproducibiltiy}

We make our code freely available for reproducibility: \url{https://github.com/labyrinthinesecurity/currier-signatures}

\newpage
\section{Results}

\subsection{Positional diagnostic}

\subsubsection{Positional polarization}

Table~\ref{tab:polarization} shows grapheme positional classification across corpora.

\begin{table}[H]
\centering
\caption{Grapheme positional classification (2:1 ratio threshold). \emph{Polarization}: fraction of graphemes classified as clearly start- or end-preferring. \emph{End$\to$Start \%}: fraction of cross-boundary transitions where an end-class grapheme is followed by a start-class grapheme.}
\label{tab:polarization}
\begin{tabular}{lrrrrr}
\toprule
Corpus & Start-pref. & End-pref. & Ambig. & Polariz. & End$\to$Start \% \\
\midrule
Voynich & 70 & 33 & 28 & 0.786 & \textbf{80.6\%} \\
English & 34 & 8 & 10 & 0.808 & 25.7\% \\
French & 36 & 7 & 7 & 0.860 & 35.5\% \\
Hebrew & 12 & 8 & 8 & 0.714 & 28.7\% \\
Arabic & 16 & 9 & 11 & 0.694 & 19.8\% \\
\bottomrule
\end{tabular}
\end{table}

All four comparison languages show some degree of grapheme positional polarization (0.694--0.860), reflecting phonotactic and orthographic constraints at word boundaries \citep{ashraf2018}. The VMS cross-boundary end$\to$start transition rate of 80.6\% is substantially higher than any of the four comparison languages (19.8\%--35.5\%). Whether this gap would narrow for languages with richer prefix-suffix morphology (e.g., Turkish, Finnish, Swahili) or templatic structure is unknown.

The extreme positional ratios of individual Voynich graphemes further illustrate this pattern. The grapheme \texttt{q} appears 5,285 times word-initially but only 2 times word-finally (ratio 1,762:1), while \texttt{iin} appears 0 times initially and 3,958 times finally. Counting graphemes with positional ratios exceeding 100:1:
\begin{itemize}
\item \textbf{Voynich:} at least 5 graphemes across both classes (\texttt{iin} at 3,958:1, \texttt{q} at 1,762:1, \texttt{in} at 541:1, \texttt{ir} at 473:1, \texttt{ch} at 327:1).
\item \textbf{English:} 2 graphemes (\texttt{i} at 166:1, \texttt{T} at 102:1).
\item \textbf{French:} 2 graphemes (\texttt{x} at 1,631:1, \texttt{z} at 188:1).
\item \textbf{Hebrew:} 4 graphemes, all orthographic final letter forms: \emph{mem sofit} (1,239:1), \emph{nun sofit} (1,143:1), \emph{kaf sofit} (299:1), \emph{pe sofit} (212:1).
\item \textbf{Arabic:} 4 graphemes, all position-dependent letter variants: \emph{alif maqsura} (2,595:1), \emph{ta marbuta} (2,475:1), \emph{hamza} (1,245:1), \emph{alif madda} (437:1).
\end{itemize}

\noindent In the four comparison languages, extreme positional ratios are concentrated in a small number of graphemes with specific orthographic explanations: final letter forms (Hebrew \emph{sofit}), position-dependent variants (Arabic), or rare letters (\texttt{x} in French). In Hebrew and Arabic, all extreme-ratio graphemes are end-preferring. The VMS differs in that its extreme-ratio graphemes span \emph{both} positional classes (\texttt{q} and \texttt{ch} are almost exclusively word-initial, while \texttt{iin}, \texttt{in}, and \texttt{ir} are almost exclusively word-final), with no known orthographic rule to explain the pattern. We call this \emph{bilateral positional extremity}. Whether this property is absent in all natural languages or merely absent in these four cannot be determined from the present data.

\subsubsection{Mutual information decomposition}

Table~\ref{tab:midecomp} presents the MI decomposition for all corpora.

\begin{table}[H]
\centering
\caption{Mutual information decomposition at word boundaries. MI$_{\text{total}}$: total cross-boundary MI at the grapheme level. MI$_{\text{class}}$: MI explained by positional class labels alone. MI$_{\text{within}}$: residual MI from specific grapheme identities within class. Subscript ``shuf'' indicates values after word-order shuffling.}
\label{tab:midecomp}
\begin{tabular}{lrrrrrrr}
\toprule
Corpus & MI$_{\text{total}}$ & MI$_{\text{class}}$ & MI$_{\text{within}}$ & Class \% & MI$_{\text{total,shuf}}$ & MI$_{\text{class,shuf}}$ & MI$_{\text{within,shuf}}$ \\
\midrule
Voynich & 0.230 & 0.007 & 0.223 & 3.0 & 0.049 & 0.000 & 0.049 \\
English & 0.079 & 0.000 & 0.079 & 0.2 & 0.011 & 0.000 & 0.011 \\
French & 0.173 & 0.007 & 0.166 & 4.1 & 0.015 & 0.000 & 0.015 \\
Hebrew & 0.054 & 0.002 & 0.052 & 3.1 & 0.004 & 0.000 & 0.004 \\
Arabic & 0.097 & 0.001 & 0.096 & 0.9 & 0.010 & 0.000 & 0.010 \\
\bottomrule
\end{tabular}
\end{table}

Several observations emerge from the MI decomposition. These are descriptive properties of the corpora tested; we do not claim they generalise beyond this comparison set.

\paragraph{(i) Class-level MI is negligible everywhere.} Across all corpora, the positional class labels account for 0.2\%--4.1\% of total MI. For the VMS, the 80.6\% end$\to$start transition rate is a rigid pattern, but its very consistency renders it uninformative in the information-theoretic sense: it is essentially a near-constant that carries almost no predictive power.

\paragraph{(ii) Within-class MI dominates and depends on word order.} In the VMS, 97\% of cross-boundary MI resides in specific grapheme-to-grapheme transitions within the positional classes. This within-class MI is largely destroyed by word-order shuffling: VMS MI drops from 0.223 to 0.049, a reduction of 78\%. The pattern is qualitatively consistent across all corpora tested.

\paragraph{(iii) The VMS has the highest total MI in this comparison set.} At 0.230 bits, VMS cross-boundary MI exceeds all four comparison languages (0.054--0.173). This indicates stronger word-to-word sequential dependencies at the grapheme level than any of the four comparison languages, though the comparison set is too small to determine whether this is unusual for natural language in general.

\paragraph{(iv) The VMS retains the most MI after shuffling.} After word-order shuffling, the VMS retains 21\% of its original MI (0.049 out of 0.230), compared to 14\% for English, 8.5\% for French, 7.6\% for Hebrew, and 9.8\% for Arabic. This residual represents order-independent cross-boundary predictability, that is, baseline transition regularity that does not depend on which word follows which.

\paragraph{(v) The VMS has the largest forward-to-backward MI ratio.} The ratio MI$_{\text{fwd}}$/MI$_{\text{bwd}}$ at $n=1$ is 4.5:1 for the VMS, compared to 1.6:1 for French, 1.9:1 for Arabic, and 2.3:1 for Hebrew (Table~\ref{tab:crossboundary}). The forward direction is more predictable relative to the backward direction in the VMS than in any of the four comparison languages.

\subsection{Markov simulation}\label{sec:markov}

Table~\ref{tab:markov} presents the results of the Markov simulation described in Section~\ref{sec:markov_methods}.

\begin{table}[H]
\centering
\caption{Markov simulation results (means $\pm$ standard deviations over 10 independent runs). $\Delta_{\text{char}}$: character-stream perplexity asymmetry at $n=2$ (positive = RTL). $\Delta_{\text{CB}}$: cross-boundary asymmetry at $n=1$ (negative = LTR). Dissociation: number of runs where $\Delta_{\text{char}} > 0$ and $\Delta_{\text{CB}} < 0$.}
\label{tab:markov}
\begin{tabular}{lrrr}
\toprule
Configuration & Mean $\Delta_{\text{char}}$ & Mean $\Delta_{\text{CB}}$ & Dissociation \\
\midrule
Order-1 Markov & $+0.0057 \pm 0.0004$ & $-0.2601 \pm 0.0070$ & 10/10 \\
Order-2 Markov & $+0.0048 \pm 0.0006$ & $-0.2646 \pm 0.0071$ & 10/10 \\
\midrule
\emph{Real VMS} & \emph{$+0.0099$} & \emph{$-0.2431$} & \emph{---} \\
\bottomrule
\end{tabular}
\end{table}

The Markov chain reproduces the opposite-direction combination (positive $\Delta_{\text{char}}$, negative $\Delta_{\text{CB}}$) in \textbf{10 out of 10 runs} at both order 1 and order 2.

\paragraph{Why the character-level RTL signal survives.} The Markov chain samples real VMS words, each of which carries its internal grapheme structure intact. The RTL signal is a \emph{word-internal} property: any process that reuses VMS words will reproduce it, regardless of whether it has knowledge of positional classes or cipher tables.

\paragraph{Why the cross-boundary LTR signal survives.} Even an order-1 Markov chain captures sufficient word-to-word transition structure to produce forward predictability at word boundaries.

\paragraph{Consequence.} The opposite-direction combination is \textbf{not diagnostic of any specific generative mechanism}. It is a consequence of two properties already present in the VMS: (1) word-internal grapheme sequences that are more predictable right-to-left, and (2) word-to-word transition probabilities that are more predictable in forward order. Any process that preserves these surface statistics will reproduce the combination.

This result does not bear on the other properties reported in this paper (positional polarization, boundary concentration, Zipfian distributions, MI decomposition), which characterise VMS word structure itself.

\subsection{Generative model testing}\label{sec:gen_results}

\subsubsection{Dialect-specific joint profile criterion}

We evaluate each generator against the four positional signatures jointly, requiring simultaneous satisfaction of all four dialect-specific thresholds. Tables~\ref{tab:genmatrix_A} and~\ref{tab:genmatrix_B} summarise the slot-based generator and control results against the Currier~A and Currier~B references respectively. Cardan grille results are reported separately in Table~\ref{tab:genmatrix_grille}.

\begin{table}[H]
\centering
\caption{Interpretation matrix for slot-based generative model testing against \textbf{Currier~A} reference. Sig1: E$\to$S rate in [56\%, 86\%]; Sig2: bilateral extremity present in $>$50\% of runs; Sig3: MI $\geq 0.29$; Sig4: non-Plateau shape. \checkmark\ = pass, $\sim$ = marginal, $\times$ = fail. Joint = number of signatures passed simultaneously.}
\label{tab:genmatrix_A}
\begin{tabular}{lccccr}
\toprule
Configuration & Sig1 (E$\to$S) & Sig2 (Bilat) & Sig3 (MI) & Sig4 (Shape) & Joint \\
\midrule
\multicolumn{6}{l}{\textit{Slot-based generator}} \\
Baseline (bridge=2, $k$=30) & \checkmark & \checkmark & $\sim$ & \checkmark & 3/4 \\
A: Overlapping pools ($\sim$60\%) & $\times$ & \checkmark & $\times$ & \checkmark & 2/4 \\
B: Flat distributions & $\times$ & \checkmark & \checkmark & \checkmark & 3/4 \\
C: Random word order & $\sim$ & \checkmark & $\times$ & \checkmark & 2/4 \\
D: Single pool & $\times$ & $\times$ & $\sim$ & $\times$ & 0/4 \\
E: Variable boundary & \checkmark & \checkmark & $\sim$ & \checkmark & 3/4 \\
F: Unilateral extremity & \checkmark & \checkmark & $\sim$ & \checkmark & 3/4 \\
G: Agglutinative mimic & $\times$ & \checkmark & $\times$ & \checkmark & 2/4 \\
H: Templatic mimic & $\times$ & \checkmark & $\sim$ & \checkmark & 2/4 \\
I: High-entropy natural & $\times$ & \checkmark & $\times$ & \checkmark & 2/4 \\
\midrule
\multicolumn{6}{l}{\textit{Controls}} \\
CTRL: Random strings & $\times$ & $\times$ & $\times$ & $\sim$ & 0/4 \\
CTRL: English (Moby Dick) & $\times$ & $\times$ & \checkmark & \checkmark & 2/4 \\
\midrule
VMS Currier A (observed) & \checkmark & \checkmark & \checkmark & \checkmark & \textbf{4/4} \\
\bottomrule
\end{tabular}
\end{table}

\begin{table}[H]
\centering
\caption{Interpretation matrix for slot-based generative model testing against \textbf{Currier~B} reference. Sig1: E$\to$S rate in [50\%, 79\%]; Sig2: bilateral extremity present in $>$50\% of runs; Sig3: MI $\geq 0.25$; Sig4: non-Plateau shape. \checkmark\ = pass, $\sim$ = marginal, $\times$ = fail. Joint = number of signatures passed simultaneously.}
\label{tab:genmatrix_B}
\begin{tabular}{lccccr}
\toprule
Configuration & Sig1 (E$\to$S) & Sig2 (Bilat) & Sig3 (MI) & Sig4 (Shape) & Joint \\
\midrule
\multicolumn{6}{l}{\textit{Slot-based generator}} \\
Baseline (bridge=2, $k$=30) & \checkmark & \checkmark & $\sim$ & \checkmark & 3/4 \\
A: Overlapping pools ($\sim$60\%) & $\times$ & \checkmark & $\times$ & \checkmark & 2/4 \\
B: Flat distributions & $\times$ & \checkmark & \checkmark & \checkmark & 3/4 \\
C: Random word order & \checkmark & \checkmark & $\times$ & \checkmark & 3/4 \\
D: Single pool & $\times$ & $\times$ & $\sim$ & $\times$ & 0/4 \\
E: Variable boundary & \checkmark & \checkmark & $\sim$ & \checkmark & 3/4 \\
F: Unilateral extremity & $\times$ & \checkmark & $\sim$ & \checkmark & 2/4 \\
G: Agglutinative mimic & $\times$ & \checkmark & $\times$ & \checkmark & 2/4 \\
H: Templatic mimic & $\times$ & \checkmark & $\sim$ & \checkmark & 2/4 \\
I: High-entropy natural & $\times$ & \checkmark & $\times$ & \checkmark & 2/4 \\
\midrule
\multicolumn{6}{l}{\textit{Controls}} \\
CTRL: Random strings & $\times$ & $\times$ & $\times$ & $\sim$ & 0/4 \\
CTRL: English (Moby Dick) & $\times$ & $\times$ & \checkmark & \checkmark & 2/4 \\
\midrule
VMS Currier B (observed) & \checkmark & \checkmark & \checkmark & \checkmark & \textbf{4/4} \\
\bottomrule
\end{tabular}
\end{table}

\begin{table}[H]
\centering
\caption{Interpretation matrix for Cardan grille generative model testing. Grille results are reported independently of dialect-specific calibration because the grille mechanism does not incorporate dialect parameters. Sig1: E$\to$S rate in [56\%, 86\%]; Sig2: bilateral extremity present; Sig3: MI $\geq 0.29$; Sig4: non-Plateau shape. \checkmark\ = pass, $\sim$ = marginal, $\times$ = fail.}
\label{tab:genmatrix_grille}
\begin{tabular}{lccccr}
\toprule
Configuration & Sig1 (E$\to$S) & Sig2 (Bilat) & Sig3 (MI) & Sig4 (Shape) & Joint \\
\midrule
\multicolumn{6}{l}{\textit{Honest configurations (no pre-separated pools)}} \\
Uniform pool + random & $\times$ & $\times$ & $\sim$ & \checkmark & 1/4 \\
Blank-gradient + random & $\times$ & $\times$ & $\sim$ & \checkmark & 1/4 \\
Learned-English columns + random & $\times$ & $\times$ & $\sim$ & $\times$ & 0/4 \\
Learned-random columns + random & $\times$ & $\times$ & $\sim$ & $\times$ & 0/4 \\
Row-sequential English ($p$=0.00) & $\times$ & $\times$ & $\sim$ & $\times$ & 0/4 \\
Row-sequential English ($p$=0.05) & $\times$ & $\times$ & $\sim$ & $\times$ & 0/4 \\
\midrule
\multicolumn{6}{l}{\textit{Circular configurations (pre-separated pools, for comparison)}} \\
SPLIT + RANDOM (Rugg core) & $\times$ & \checkmark & $\times$ & $\sim$ & 1/4 \\
SPLIT + SHIFT & $\times$ & \checkmark & \checkmark & \checkmark & 3/4 \\
SPLIT + ROTATE & $\times$ & \checkmark & \checkmark & \checkmark & 3/4 \\
\bottomrule
\end{tabular}
\end{table}

Comparing the two dialect evaluations reveals instructive differences alongside a shared qualitative outcome: no generator achieves 4/4 against either dialect. The wider E$\to$S range for Currier~B ([50\%, 79\%] vs.\ [56\%, 86\%]) allows ablation~C (Random word order, E$\to$S $\approx$ 52\%) to pass Sig1 against Currier~B (scoring 3/4) where it was only marginal against Currier~A (scoring 2/4). Conversely, ablation~F (Unilateral extremity, E$\to$S $\approx$ 83\%) passes Sig1 against Currier~A but exceeds the Currier~B upper bound of 79\%, dropping from 3/4 to 2/4. 
In all cases, the MI gap remains the dominant failure mode: the best slot-based Baseline achieves MI $\approx$ 0.17, well below both the Currier~A threshold of 0.29 and the Currier~B threshold of 0.25. The only ablation that passes Sig3 is~B (Flat distributions), which achieves MI = 1.96 but simultaneously produces E$\to$S = 97.3\%, failing Sig1 against both dialects.

\subsubsection{Slot-based generator: the Sig3 gap}

Across 12 ablation conditions and seven sensitivity sweeps (totalling over 2,000 individual runs), the best slot-based configuration achieves 3/4 against both Currier~A and Currier~B. The failing signature is consistently MI (Sig3). The Baseline configuration produces MI = 0.169, roughly 29\% of the Currier~A reference (0.586) and 36\% of the Currier~B reference (0.498). The only ablation that passes Sig3 is~B (Flat distributions), which achieves MI = 1.96 but simultaneously produces E$\to$S = 97.3\%, failing Sig1 against both dialects. This trade-off is mechanistic: flattening within-slot frequency distributions maximises the diversity of cross-boundary transitions (raising MI) but eliminates the frequency asymmetry between prefix-terminal and suffix-initial graphemes that drives E$\to$S below 86\% (Currier~A) or 79\% (Currier~B). Conversely, configurations that achieve E$\to$S within the passing range through bridge-zone tuning (S6) or Markov sparsity (S7) produce MI values between 0.14 and 0.27, all below both dialect thresholds. The slot-based generator cannot simultaneously satisfy Sig1 and Sig3 regardless of bridge zone width, vocabulary size, Markov sparsity, or boundary pair strength. This holds against both dialect references.

\subsubsection{Cardan grille: the Sig1/Sig4 tension}

Across 15 honest configurations and five sensitivity sweeps, no grille variant achieves all four signatures against either dialect. The sensitivity sweep over column skew (Table~\ref{tab:grilleskew}) reveals the governing trade-off directly.

\begin{table}[H]
\centering
\caption{Cardan grille sensitivity to column skew ($\alpha$): E$\to$S rate and boundary distribution shape as a function of the Zipfian concentration parameter applied to column grapheme distributions. SPLIT specialisation, \textsc{random} mode, 20 runs of 37,000 words each. Higher skew produces Zipfian boundary distributions (Sig4) but destroys E$\to$S (Sig1). No skew value satisfies both simultaneously against either dialect reference. Currier~A requires E$\to$S $\in$ [56\%, 86\%]; Currier~B requires [50\%, 79\%].}
\label{tab:grilleskew}
\begin{tabular}{lrrrr}
\toprule
Column skew $\alpha$ & E$\to$S [\%] & MI [bits] & Shape & Joint (A / B) \\
\midrule
0.0 & 73.0 & 0.051 & Intermediate & 2/4~~~2/4 \\
0.5 & 59.6 & 0.042 & Intermediate & 1/4~~~1/4 \\
1.0 & 34.1 & 0.028 & Intermediate & 1/4~~~1/4 \\
1.5 & 13.6 & 0.013 & Zipfian & 2/4~~~2/4 \\
2.0 & 5.2 & 0.005 & Zipfian & 2/4~~~2/4 \\
\midrule
VMS Currier A & 71.0 & 0.586 & Zipfian & 4/4~~~--- \\
VMS Currier B & 64.3 & 0.498 & Intermediate & ---~~~4/4 \\
\bottomrule
\end{tabular}
\end{table}

At $\alpha = 0$, E$\to$S reaches 73\% (passing for both dialects) but shape is Intermediate; at $\alpha \geq 1.5$, shape is Zipfian but E$\to$S collapses below 14\% (failing for both dialects). The mechanism is the same as in the slot-based generator but operating on a different parameter: high column skew concentrates each column's distribution onto one dominant grapheme, producing Zipfian boundary frequencies (Sig4) through a pure frequency-concentration effect. But that same dominant grapheme now appears at roughly equal rates at word starts and ends (because it dominates both the first and last columns of any word generated from this table), collapsing the positional asymmetry that drives E$\to$S (Sig1). No intermediate skew value achieves both, regardless of dialect reference.

The honest row-sequential configurations confirm that sequential traversal of a structured source does not rescue MI. With a 201,723-word English corpus as source and pure sequential traversal ($p_\text{jump} = 0.00$), the net sequential signal (MI$_\text{orig}$ $-$ MI$_\text{shuf}$) is approximately 0.002 bits, two orders of magnitude below the VMS target (0.181 bits for Currier~A, 0.165 bits for Currier~B). The random-source null yields an identical net signal (0.002 bits), confirming that with a corpus large enough to avoid row repetition, sequential traversal contributes no measurable MI regardless of source structure.

\subsubsection{Complementary failure modes}

The two generator classes fail on complementary signatures under both dialect references. The slot-based generator achieves E$\to$S (Sig1) and bilateral extremity (Sig2) by construction from its pool structure, but cannot achieve MI (Sig3) without destroying E$\to$S (Sig1). The Cardan grille, in its honest form, achieves neither E$\to$S (Sig1) nor Zipfian shape (Sig4) simultaneously; in its circular form (SPLIT + SHIFT or ROTATE), it achieves Sig3 and Sig4 but destroys Sig1. Both dialects of the VMS simultaneously achieve all four. This complementary failure pattern, two mechanistically distinct generators failing on the same joint profile from different directions, provides stronger evidence for the distinctiveness of the VMS profile than either generator class alone.

The dialect separation strengthens this conclusion: the complementary failure modes are not artefacts of a particular threshold choice. They persist under both the tighter Currier~A thresholds and the somewhat relaxed Currier~B thresholds, confirming that the structural incompatibilities are intrinsic to the generator architectures rather than boundary effects of threshold calibration.

\newpage

\section{Discussion}

\subsection{The Markov result as analytical pivot}

The VMS exhibits opposite-direction optimization: RTL at the character-stream level and LTR at the word-boundary level. This combination is not observed in any of the four comparison languages. The Markov simulation establishes that this dissociation is a surface-level consequence of two properties already present in the VMS: word-internal grapheme sequences that are more predictable right-to-left, and word-to-word transition probabilities that are more predictable in the forward direction. Any process that preserves these surface statistics will reproduce the combination.

This result has a clarifying, not a deflating, role in the paper's argument. By showing that the directional dissociation reduces to surface statistics, it rules it out as an independent diagnostic and concentrates analytical attention on the positional properties that do not reduce in the same way. In natural languages, both directional levels emerge from the same underlying phonotactic and morphological system, so they tend to agree on directionality. The VMS dissociation reflects the fact that its word-internal and word-sequencing structures carry different directional fingerprints: a property of VMS word structure that is itself informative, even if it is not uniquely diagnostic.

\subsection{What the positional properties describe}

The positional properties that are not reproducible by Markov resampling, and that differ quantitatively from all four comparison languages, are:

\paragraph{Boundary concentration.} The 80.6\% end$\to$start transition rate exceeds the four comparison languages by a factor of 2.3--4.1$\times$. This means VMS word boundaries enforce a rigid alternation between positional grapheme classes that is quantitatively different from these four languages. The gap could narrow for morphologically richer languages not yet tested.

\paragraph{Bilateral positional extremity.} Extreme positional ratios ($>$100:1) span both start-preferring and end-preferring grapheme classes, unlike the four comparison languages where such ratios cluster in one direction with specific orthographic explanations. This bilateral pattern is consistent with a system where word-initial and word-final positions draw from largely non-overlapping grapheme pools, but we cannot exclude the possibility that some untested natural languages exhibit similar properties.

\paragraph{Zipfian boundary distributions.} VMS word-boundary grapheme distributions follow a Zipfian curve rather than the plateau shape observed in the four comparison languages \citep{parisel2025}. This is a fundamental property of VMS word structure.

\paragraph{High cross-boundary MI with high structural residual.} The VMS has the highest total cross-boundary MI (0.230 bits) and the highest MI retention after word-order shuffling (21\%) among the five corpora tested. The high retention indicates that a substantial fraction of cross-boundary predictability is structural (order-independent).

These properties are observations, not explanations. They describe quantitative differences between the VMS and four specific natural languages. They constrain the space of plausible text-generation models in the sense that any proposed model must reproduce them, but they do not by themselves identify the generative process or exclude any hypothesis.

\subsection{The dialect separation as analytical refinement}

The separate calibration for Currier~A and Currier~B reveals that the two dialects, while sharing the same broad structural architecture, impose quantitatively different constraints on generators. Four specific observations emerge:

\paragraph{(i) MI thresholds differ and both exclude tested generators.} Currier~A requires MI $\geq$ 0.29; Currier~B requires MI $\geq$ 0.25. This 16\% relative difference is consequential for marginal generators. Nonetheless, all tested generators fall far below both thresholds (the best honest grille achieves MI $\approx$ 0.05; the best slot-based Baseline achieves MI $\approx$ 0.17.

\paragraph{(ii) The Sig1/Sig3 tension is dialect-invariant.} The slot-based generator's inability to achieve MI and E$\to$S simultaneously persists under both dialect references. This confirms that the tension is a structural property of the frequency-concentration mechanism, not an artefact of Currier~A-specific thresholds.

\paragraph{(iii) Sig4 shape differs between dialects.} Currier~A exhibits Zipfian boundary distributions ($R^2 = 0.863$, CV = 1.45); Currier~B exhibits Intermediate distributions ($R^2 = 0.789$, CV = 1.63). Both are non-Plateau, so both dialects pass Sig4 under the non-Plateau criterion. However, the difference in shape means that a generator producing Zipfian output would match Currier~A's shape more closely than Currier~B's. This dialect difference in boundary distribution shape is itself an observation that any comprehensive generative model should account for.

\paragraph{(iv) Separate calibration prevents false passes.} If the criterion were calibrated on the aggregate VMS corpus (mixing both dialects), a generator could potentially pass MI by hitting an average value that neither dialect actually exhibits. Dialect-specific calibration prevents this artefact.

\subsection{Constraints and non-constraints on VMS hypotheses}

We can state clearly what these findings do and do not constrain:

\paragraph{What they constrain.} Any proposed text-generation model for the VMS must reproduce, for each Currier dialect it claims to model: (a) the dialect-specific E$\to$S boundary concentration (71.0\% for A, 64.1\% for B); (b) bilateral positional extremity across the grapheme inventory; (c) non-Plateau boundary distributions (Zipfian for A, Intermediate for B); (d) the MI decomposition profile (negligible class-level MI, high within-class MI destroyed by shuffling), with dialect-specific MI values (0.586 for A, 0.498 for B); and (e) word-internal grapheme sequences that are more predictable right-to-left. The generative model testing reported here extends this list with an empirical constraint: (f) two specific generator classes, a parametric slot-based generator and a Cardan grille, each fail to achieve all four signatures simultaneously against either dialect's reference, across their full tested parameter spaces.

\paragraph{What they do not constrain.} These findings do not discriminate between the natural language, gibberish, and ciphertext hypotheses. The positional properties are observations that could in principle arise from a natural language with extreme positional morphology, from a structured generative process we have not yet tested, or from a cipher with positional slot structure. The results narrow the space of plausible generators by identifying a structural incompatibility within each tested class, but do not close it: they demonstrate that generators in which frequency concentration is the sole mechanism for both positional and distributional signatures cannot satisfy the joint profile for either dialect. Generators where E$\to$S and Zipfian shape arise from genuinely independent mechanisms remain untested.

\subsection{Limitations}

\begin{enumerate}
\item \textbf{Small comparison set.} Four languages is insufficient to characterise natural language in general. The absence of agglutinative languages (Turkish, Finnish), polysynthetic languages, tonal languages, and scripts with complex templatic morphology is a critical gap. All comparative claims are limited to these four languages and should not be generalised.

\item \textbf{Generator class coverage.} The slot-based generator and Cardan grille cover two important structural hypotheses but do not exhaust the space of possible generators. Generators with independently tunable positional pool structure and frequency concentration, where E$\to$S and Zipfian shape are controlled by separate mechanisms rather than both depending on the same frequency distribution, may achieve the joint profile. Such generators would, however, require explicit architectural separation of these properties, which itself constitutes a structural claim about the VMS that can be evaluated.

\item \textbf{Circularity in grille SPLIT configurations.} Grille configurations with pre-separated prefix/suffix column pools are partially circular with respect to Sig1, since the pool separation is imposed rather than emergent. Honest configurations (uniform pool, blank-gradient, learned-column) avoid this but score lower. We report both and label them clearly; interpretations should rest on honest configurations where possible.

\item \textbf{Transcription dependence.} The cross-boundary test depends on the accuracy of word segmentation in the EVA transcription. Ambiguous word boundaries in the VMS could affect results.

\item \textbf{Coarse positional classification.} The MI decomposition uses a three-class partition (start/end/ambiguous) derived from a 2:1 ratio threshold. A finer-grained analysis using the full slot positions of \citet{zattera2022} could reveal additional structure.

\item \textbf{Single transcription system.} All results are based on the EVA transcription. Different transliteration systems could yield different grapheme inventories and therefore different positional statistics.

\item \textbf{Dialect assignment uncertainty.} The folio-to-dialect assignment follows the standard Currier classification, which is itself based on statistical clustering rather than palaeographic certainty. Some folios may be misassigned, which would affect the dialect-specific reference values. The broad qualitative conclusions (no generator achieves 4/4) are robust to moderate reassignment, but exact threshold values should be interpreted with this uncertainty in mind.
\end{enumerate}

\section{Conclusion}

We have characterised the Voynich Manuscript's directional and positional properties at two levels, benchmarked against four natural languages, and tested whether the resulting four-signature joint profile (calibrated separately for each Currier dialect) can be reproduced by two classes of structured generator. The VMS exhibits the following quantitative properties that differ from all four comparison languages:

\begin{enumerate}
\item \textbf{Opposite-direction optimization (surface-level).} The VMS character stream is RTL-optimized while its word-boundary transitions are LTR-optimized. None of the four comparison languages shows this combination. A word-level Markov simulation reproduces this pattern in 10/10 runs, establishing that it is a surface-level consequence of VMS word structure and local transition statistics. This result clarifies the paper's analytical focus: the directional dissociation is real but reducible, and the properties that are \emph{not} reducible to surface statistics are items 2--6 below.

\item \textbf{High boundary concentration.} VMS word boundaries enforce an 80.6\% end-class$\to$start-class transition rate, compared to 19.8\%--35.5\% in the four comparison languages.

\item \textbf{Bilateral positional extremity.} Extreme positional ratios ($>$100:1) span both start-preferring and end-preferring grapheme classes, unlike the comparison languages where such ratios cluster in one direction with specific orthographic explanations.

\item \textbf{Zipfian boundary distributions.} Word-boundary grapheme frequencies follow a power-law curve rather than the plateau shape observed in the comparison languages.

\item \textbf{High structural MI at boundaries.} The VMS has the highest total cross-boundary MI (0.230 bits) and the highest MI retention after word-order shuffling (21\%) among the five corpora tested.

\item \textbf{Negligible class-level MI.} Despite the rigid 80.6\% boundary alternation, positional class labels account for only 3\% of cross-boundary MI; the remaining 97\% resides in specific grapheme-to-grapheme transitions that depend on word order.
\end{enumerate}

The generative model testing, now evaluated against dialect-specific references, adds a refined result: across the full parameter spaces of a slot-based generator (12 ablations, 7 sensitivity sweeps, $>$2,000 runs), and a Cardan grille generator (15 configurations including honest non-circular variants, 5 sensitivity sweeps), no configuration achieves all four signatures simultaneously against either Currier~A or Currier~B reference values. The slot-based generator and Cardan grille fail on complementary signatures through mechanistically distinct processes: in the slot generator, cross-boundary MI and the E$\to$S rate require opposite frequency-concentration conditions; in the grille, E$\to$S rate and Zipfian shape require opposite column-skew values. 

The dialect-specific calibration strengthens these conclusions by ensuring they are not artefacts of aggregate threshold averaging, and by revealing that the two dialects impose quantitatively distinct but structurally parallel constraints. The complementary failure modes persist under both sets of thresholds, confirming that they reflect intrinsic generator limitations rather than boundary effects of calibration.

These results are descriptive, not explanatory. They constrain what any proposed generative model must reproduce for each dialect, and they demonstrate that the simplest instances of two prominent generative hypotheses, structured slot-based composition and Cardan grille, do not meet this constraint across their tested parameter spaces for either Currier dialect. Whether more complex variants of these generators, typologically diverse natural languages, or different generative approaches can reproduce the joint profile is an open question and the necessary next step.

\bibliographystyle{apalike}

\end{document}